%% file: springer_main.tex
\documentclass{springer_style/svproc}
\usepackage{cite}
\usepackage{amsmath,amssymb,amsfonts}
\usepackage{algorithmic}
\usepackage{graphicx}
\usepackage{textcomp}
\usepackage{xcolor}
\begin{document}
\title{Training Heterogeneous Features in Sequence to Sequence Tasks: Latent Enhanced Multi-filter Seq2Seq Model}
\titlerunning{Latent Enhanced Multi-filter Seq2Seq Model}
%
\authorrunning{Y. Yang et al.}
\author{Yunhao Yang \inst{1} \and
Zhaokun Xue\inst{2}}
%
%
\institute{University of Texas at Austin, Austin TX 78705, USA \\
\email{yunhaoyang234@utexas.edu}\\
 \and
University of Texas at Austin, Austin TX 78705, USA \\
\email{xuezhaokun@utexas.edu}}
\maketitle              
\input{tex/abstract}

\input{tex/introduction}
\input{tex/related_work}
\input{tex/method}
\input{tex/experiments}
\input{tex/discussion}
\input{tex/conclusion}
%
%
\bibliographystyle{springer_style/splncs04}
\bibliography{ref}

\end{document}

%% file: tex/abstract.tex
\begin{abstract}
In language processing, training data with extremely large variance may lead to difficulty of language model's convergence. It is difficult for the network parameters to adapt sentences with largely varied semantics or grammatical structures. To resolve this problem, we introduce a model that concentrates the each of the heterogeneous features in the input sentences. Build upon the encoder-decoder architecture, we design a latent-enhanced multi-filter seq2seq model (LEMS) that analyzes the input representations by introducing a latent space transformation and clustering. The representations are extracted from the final hidden state of the encoder and lie in the latent space. A latent space transformation is applied for enhancing the quality of the representations. Thus the clustering algorithm can easily separate samples based on the features of these representations. Multiple filters are trained by the features from their corresponding clusters, the heterogeneity of the training data can be resolved accordingly. We conduct two sets of comparative experiments on semantic parsing and machine translation, using the Geo-query dataset and Multi30k English-French to demonstrate the enhancement our model has made respectively.

\keywords{Representation Learning, Recurrent Neural Network, Latent Subspace, Neural Machine Translation, Semantic Parsing}
\end{abstract}

%% file: tex/introduction.tex
\section{Introduction}

Semantic parsing is the process of transforming a natural language utterance to a formal representation of its meaning, of which the machine can understand. It is widely applied in the field of natural language processing.
The goal is to construct a proper mapping from the original data space to a latent space that keep the features of the original data.
A way of solving semantic parsing problems is using a sequence to sequence model.

A sequence to sequence model maps a fixed-length input with a fixed-length output where both the input and output can have variable length. It is composed by an encoder-decoder model, which is an approach that using recurrent neural networks (RNN) as the underlying component.

The architecture of an encoder-decoder model comprises of two components. The first component is the encoder, which maps a variable-length sequence to a vector in the latent space. The second component is the decoder, which maps the final internal state vector in the latent space to a variable-length sequence.

The encoder is a stack of RNN cells. Iteratively, it takes as input a single element of the input sequence and the internal state from the last RNN cell, and propagates its internal state to the next RNN cell. The final internal state vector in the latent space is the encoder's summary of the whole sequence and serves to be the input of the decoder. 
Previous research has shown that word sequences consist of similar attributes, such as semantics or grammatical structures, are topologically close to each other. We can exploit and explore the features of the word sequences in the latent space.
Similar to the encoder, the decoder is also comprised of a sequence of RNN cells. It predicts one token at a time given as input the previously predicted token and the internal state of the previous RNN cell.

Representation learning is a machine learning technique for data analyzing. The neural networks perform space transformation that maps the input data from the input space to some other spaces, such as the latent space. In this work, the latent space refers to the output space of the encoder. It is a lower-dimension space comparing to the input space that preserve the features of the input data. We project a input data record on the latent space to obtain a vector, we denote the vector as the \textsc{representation} of this input record. Due to the abilities of dimension-reduction and feature-preserving, we can extract the representations from the latent space and directly analyze them.

In machine learning, representation learning is the technique of automatically extracting representations from raw data and thus make it computationally feasible to process the data. Because different representations can differ in the explanatory factors of variation behind the data, the success of a given algorithm depend on the quality of the extracted representation \cite{bengio2013representation}. 

The representation of a sentence is extracted from the last hidden state of the encoder. We denote the last hidden space of the encoder as the latent space. Sentence representations are lying in the latent space, whose features cannot be interpreted directly. Instead, the latent space representations are machine recognizable/understandable through a quantitative spatial representation or modeling.

In this work, we use an encoder-decoder model as the technique for representation learning, train the representations of the sentences. To further enhance the representations, we embedded a multi-layer perceptron enhancer. Both the encoder-decoder model and the enhancer is designed for generating good representations of the sentences. We assume that the representations of similar sentences should be closer. Hence we can divide the data into clusters and use feature-specific decoders to decode these representations, where each decoder is only responsible for one cluster. So, our paper has two major contributions:
\begin{itemize}
    \item introduce a multi-decoder structure that can resolve the convergence problem that caused by large variance in the training data,
    \item demonstrate a positive correlation between the quality of the sentence representations and model's overall performance, which means we can further improve the model by enhancing the way of generating representations.
\end{itemize}

In our paper, we first discuss some related research in Section \ref{sec:relate}. Then, we describe the architecture of our Latent Enhanced Multi-filter Sequence to Sequence Model and the details of the training pipeline in Section \ref{sec:method}. Later in Section \ref{sec:experiment}, we conduct two sets of sequence to sequence experiments and empirically compare our results with the traditional RNN model and other benchmarks. We also provide samples to demonstrate why our multi-filter structure would enhance the performance. Last, we discuss the relationship between the quality of latent space clustering and model's performance in Section \ref{sec:discussion}, and point out some potential directions for future research.

%% file: tex/related_work.tex
\section{Related Work}
\label{sec:relate}

Many researchers are developing the networks for sequence to sequence tasks, including transformers \cite{} and recurrent encoder-decoder networks \cite{autoencoder_machine_translation}. We concentrate on the encoder-decoder structure in this work. The encoder encodes input sequences and generate the representations that keep the semantics and structures. The decoder tries to understand the representations and generate output sequences from these representations \cite{text_reconstruction}.

The encoder-decoder network is generally trained in a supervised manner\cite{Sherstinsky_2020}. Comparing with feed-forward networks which takes a fixed-length context to predict next words in a sequence, recurrent neural networks(RNN) uses their internal state to process input sequence of variable length, making it capable of taking into account all of the predecessor words \cite{dupond2019thorough}. LSTM\cite{lstm} is a special kind of RNN, and it addresses the vanishing gradient problem in traditional RNNs \cite{bengio1994learning}, where during backpropagation, gradient either vanishes or explodes exponentially. The addition of the forget gate into the cell architecture gives LSTM gain better control of the gradient value, and it is therefore relatively insensitive to gap length. Our LEMS is build on top of the LSTM structure and outperforms the Bidirectional LSTM.

In addition to using recurrence structure as the main building block of the encoder-decoder network, there has been success in the attention mechanisms. Self-attention is an attention mechanism that computes a representation of the sequence by relating different positions of a single sequence. The structure reduces the cost of modeling of dependencies between inputs of distant position to a constant number of operations, which makes computationally efficient\cite{vaswani2017attention}. The dominant sequence transaction models are based on complex recurrent neural networks in an encoder-decoder configuration. The best performing models also connect the encoder and decoder through an attention mechanism. 

In addition to ordinary supervised learning methods, some studies have also developed an autoencoder structure that exploits sequence representations in an unsupervised manner. Research has explored this approach to machine translation, where they seek to leverage language-agnostic multilingual sentence representations to easily generalize to new languages while leveraging pre-trained cross-lingual word embeddings\cite{mullov2021unsupervised}. Research also shows that latent space representations can take advantage of these features and better preserve key properties of the original data\cite{yang2021identifying}. Thus the learned representations can be used in the sequence to sequence tasks for feature analysis.

Therefore, there are several works trying to divide the latent space into subspaces, where each subspace consists of a set of homogeneous features \cite{goyal2017nonparametric,bouchacourt2018multi}. Then, the data with similar features will be located in the same subspace. Some works simply use clustering algorithms to separate the latent space \cite{yang2017towards, jabi2019deep, dilokthanakul2016deep}.

These studies show that partitioning the latent space can improve the quality of representation in some ways. Another study used this conclusion to develop a multi-filter Gaussian mixture variational autoencoder that enhances the performance in Camera ISP\cite{yang2021learning}. Following this idea, one of the previous research improves the encoder-decoder model by dividing latent space and using multiple decoders. The Multi-filter Gaussian Mixture Autoencoder\cite{yang2021representation} utilizes an autoencoder to generate the representations of the input. This model uses a Gaussian Mixture Model to divide latent space representations into clusters. Each cluster of data is decoded by a unique decoder that only concentrates on the features from this cluster.

%% file: tex/method.tex
\section{Methodology}
\label{sec:method}
\begin{figure*}[!ht]
    \begin{center}
        \includegraphics[width=0.9\linewidth]{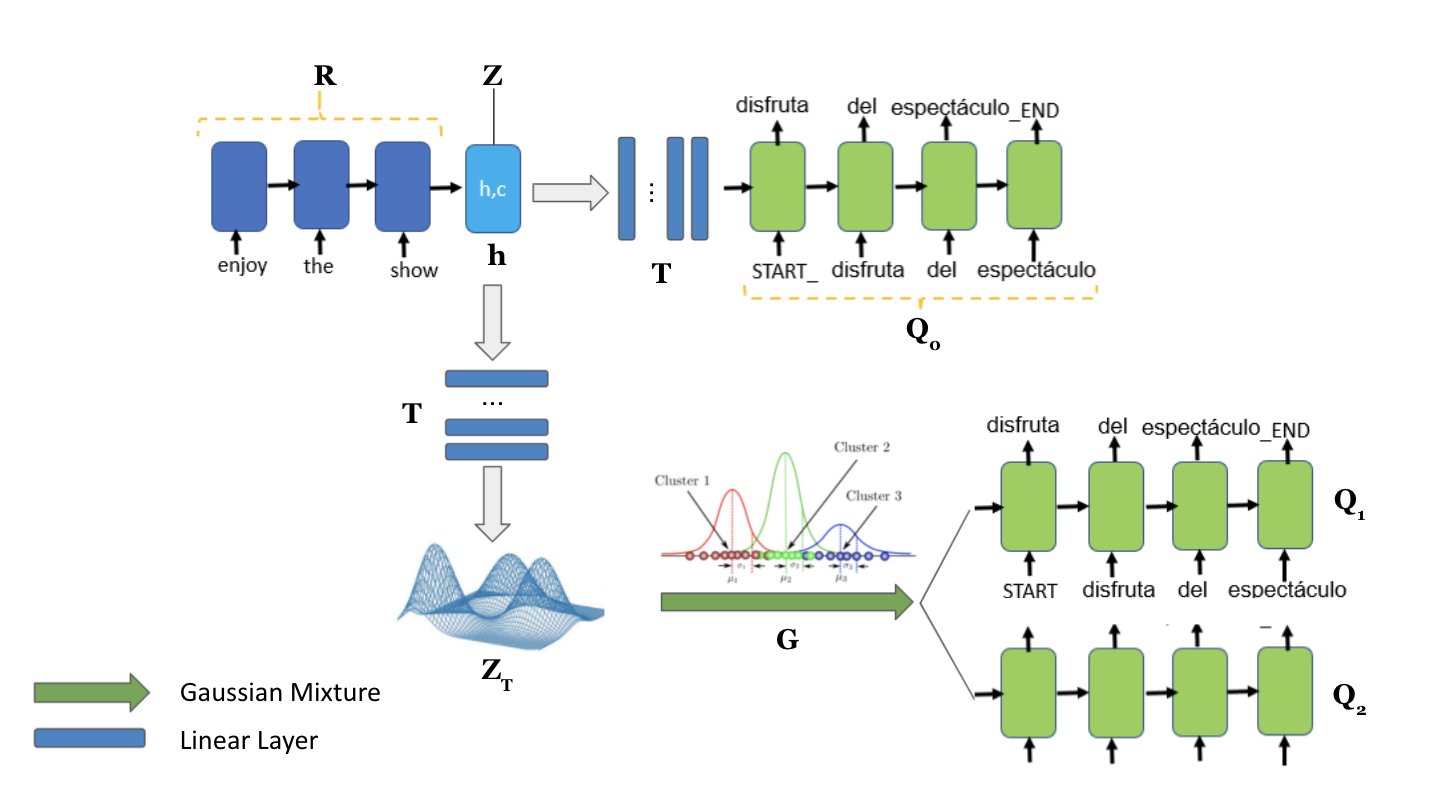}
    \end{center}
    \caption{Network Architecture: the encoder-decoder model on the top consists of an encoder $R$ and a decoder $Q_0$. The encoder-decoder model learns the representations of the input-output pairs. The representations are denoted as $h$ on the latent space $Z$. A transformation $T$ is applied to the latent space to enhance the quality of the representation vectors. The enhanced vectors are lying in a new latent space $Z_T$. A Gaussian mixture model is trained for clustering vectors in $Z_T$. Each cluster is corresponding to a filter, which is only trained and used to decode the vectors from its correspondence cluster.}
    \label{fig:architecture}
\end{figure*}

Our proposed sequence to sequence model has the following pipeline: an encoder-decoder model to learn the representations of the input-output pairs; a transformer is applied for enhancing the quality of the representations; the representations are then clustered and decoded in a defined way using several filters. The model is denoted as the Latent-Enhanced Multi-filter Sequence-to-sequence Model (LEMS). A visualization of this architecture is presented in Figure \ref{fig:architecture}. This model is evolved from Multi-filter Gaussian Mixture Autoencoder (MGMAE) \cite{yang2021representation} and we will compare the performance between these two networks in Section \ref{sec:experiment}.

\subsection{Latent Transformed Encoder-Decoder Model}
The latent transformed encoder-decoder model consists of an encoder $R$, a dummy decoder $Q_0$, and a transformer $T$ in between the encoder and decoder. We feed a sequence of words as input to the encoder step by step. The encoder cell consists of a bi-directional LSTM and a fully connected layer. Each time we obtain a hidden state from the encoder cell, then we feed the hidden state and the next word to the encoder cell. We continue the steps until all the end of the sentence. Then, we use the last hidden state $h$ as the latent space representation.

The hidden states generated by the encoder are the representations of the input-output pairs. Space where the representations lying in is denoted as the latent space $Z$. Within the latent space, a transformer is applied for enhancing the representations.
\begin{center}
    $\overline{h} = T(h)$
\end{center}

The transformer is a multi-layer perception whose input and output dimensions are the same. The enhanced representation, denoted as $\overline{h}$, is feed as the input of the decoder.

The dummy decoder $Q_0$ generates outputs from the enhanced representations. The decoder continues taking in the hidden states as input and generating output tokens, until reaching the "end of sentence" token. We construct a bi-directional LSTM connected to a fully connected layer with logarithmic softmax activation. and use them as the decoder cell. We record the output sequences from the encoder and use them to perform a dot product attention mechanism \cite{luong2015effective} in the decoding stage. In the decoding stage, we compute the attention weights associated with each token in the input sequence and multiply the weights with the current decoding hidden state.

To optimize the parameters in this encoder-decoder model, a Negative Likelihood Loss is utilized to update gradients. The loss is presented in Equation \ref{eq:nll}.

\begin{equation}
    \label{eq:nll}
    \mathcal{L}(\overline{x}, y) = \{ l_1, ..., l_N\}^T, \quad l_n = -w_{y_n} \overline{x}_{n, y_n}
\end{equation}
where $\overline{x}$ is the output tensor and $y$ is the label tensor.

The loss is used to optimize all the parameters in the encoder, decoder, as well as the transformer. After the model is fine tuned, the encoder and transformer are preserved for generate the enhanced representations. Then, we can use the Gaussian Mixture Model to divide the representations into vectors. The dummy decoder $R_0$ is only used for helping training the encoder $R$ and transformer $T$, therefore, it will not be used anymore in the future steps.

\subsection{Gaussian Mixture Model on Transformed Latent Space}
The transformed hidden vectors (representations) are projected on $Z_T$. Then, we follow the Gaussian Mixture Model (GMM) \cite{Reynolds2015} presented in MGMAE\cite{yang2021representation} to divide the enhanced latent space. The input of the GMM $G$ is the vector on $Z_T$ whose dimension is the hidden layer dimension of the encoder-decoder model. The output of $G$ is the cluster index $c_i$, where $c_i = 1,...,n$, $n$ is the total number of clusters.

We optimize the parameters of GMM using through expectation maximization, over the given training data. The goal of GMM is to divide the data into $M$ clusters and learn the mean and standard deviation for each normally distributed cluster.
\begin{equation}
\label{eq:gm}
    p(x|\lambda) = \sum_{i=1}^{M} w_i g(x|\mu_i, \Sigma_i),
\end{equation}
where $x$ is the latent space representation generated by the encoder.

We compute the densities of each Gaussian distribution following the next equation:
\begin{equation}
    g(x|\mu_i, \Sigma_i) = \frac{1}{(2\pi )^{L/2} \sqrt{|\Sigma_i|}} \exp ( \frac{d' \Sigma^{-1} d}{2} )
\end{equation}
where $d = x - \mu_i$, $\mu_i$ and $\Sigma_i$ are the mean vector and covariance matrix respectively. $L$ is the dimension of the latent space. The mean and covariance are represented by the notation $\lambda = \{ \mu_i, \Sigma_i, w_i \}$.

After the data are divided into clusters, there are multiple filters (decoders) for constructing the outputs.

\subsection{Multi-filter Decoding}
LEMS has a user-defined number of filters $Q_1, ..., Q_n$ to construct the output sequences from the enhanced representations in the space $Z_T$. Each filter shares an identical structure with the dummy decoder $Q_0$ in the encoder-decoder model.

In the training stage, we apply the GMM to assign data into $n$ clusters. Then, we construct $n$ filters (decoders) for which each filter is associated with one cluster of data. We train the filter $Q_i$ only use the data from the $i$th cluster $c_i$, so that $Q_i$ is concentrating only on the features from $c_i$. We use the same loss as presented in Equation \ref{eq:nll} to update the weights of each filter. While training the filters, we fix the weights of the encoder and the transformer $T$ and perform gradient updates for each filter independently. So that the gradients computed from $Q_i$ will not affect the weights of $Q_j$ when $i \ne j$.

In the testing stage, the input data is first encoded and projected on the latent space by the encoder and enhanced by the transformer. After that, the Gaussian mixture model classifies which cluster the input data belongs to. The sample classified into the cluster $c_i$ will go through the corresponding filter $Q_i$ for constructing the output.

%% file: tex/experiments.tex
\section{Experiments and Results}
\label{sec:experiment}
In order to demonstrate the effectiveness of our Latent-Enhanced Multi-filter Seq2Seq architecture, some comparative experiments are conducted. Our experiments try to support two claims:
\begin{itemize}
    \item the multi-filter architecture can better analyze the features and generates better outputs.
    \item the transformer $T$ for latent enhancement does actually improve the performance.
\end{itemize}

We conduct two sets of experiments on two of the well-know sequence to sequence tasks- semantic parsing and neural machine translation- to support these two claims. For each set of experiments, we train an encoder-decoder model with bi-directional LSTM as a benchmark. We provide empirical results to indicate that our multi-filter architecture and the transformer $T$ can enhance the performance compare to the benchmark.

\subsection{Semantic Parsing in Geo-query}
We first conduct a set of experiments on the Geo-query dataset\cite{10.5555/1864519.1864543} that performs semantic parsing on geographical questions. We use the token level accuracy and the denotation accuracy to evaluate the performance. The token level accuracy is based on simple token-level comparison against the reference logical form. And the denotation accuracy is the percentage of denotation match, as in \cite{Jia_2016} and \cite{liang-etal-2011-learning}.

We conduct a set of comparative experiments, hence the training and validation environments are identical. The network and training specifications are stated in Table \ref{tab:specification}.

We have trained and evaluated several models with different number of filters. So that we can determine the best number of filters to use in this task. We discover that the optimal number of filters for this dataset is 2. In addition, We use the ordinary encoder-decoder model and MGMAE\cite{yang2021representation} as the benchmarks to show the enhancement made by our LEMS. 
By comparing with MGMAE, we can also demonstrate the effectiveness of latent space enhancement by the transformer $T$.

\begin{table}[!ht]
    \label{tab:specification}
    \centering
    \caption{Experimental Specifications.}
    \begin{tabular}{||c c c c||}
     \hline
         Training Size & Validation Size & Epoch & Learning Rate \\
         480 & 120 & 10 & 0.001 \\
          \hline
         Hidden Dimension & Dropout & Embedding Dimension & Bidirectional \\
         200 & 0.2 & 150 & True\\
          \hline
    \end{tabular}
\end{table}

We perform each experiment five times and report their average in Table \ref{tab:1}. The results show that our LEMS is able to significantly outperform the other baseline methods.

\begin{table}[!htbp]
\caption{Performance comparison between the baselines and LEMS models under different number of filters.}
\begin{center}
 \begin{tabular}{||c c c||} 
 \hline
 Model & Token & Denotation \\
 \hline
 Enc-Dec & 77.4 & 43.8 \\
 SCISSOR\cite{10.5555/1706543.1706546} & 72.3 &  \\
 KRISP\cite{kate-mooney-2006-using} & 71.7 &  \\
 WASP\cite{wong-mooney-2006-learning} & 74.8 &  \\
 MGMAE\cite{yang2021representation} & 77.9 & 48.6 \\
 2-filter LEMS & 78.3 & 50.9 \\
 3-filter LEMS & 77.2 & 47.1 \\
 4-filter LEMS & 74.8 & 44.6 \\
 \hline
\end{tabular}
\end{center}
\label{tab:1}
\end{table}

By comparing with the ordinary encoder-decoder model, we can prove that the multi-filter structure with latent enhancement does significantly improve the results, in both metrics. By comparing with the MGMAE, we can observe a small improvement, which indicates the transformer $T$ does contribute to the model.

\begin{figure}[!ht]
    \centering
    \includegraphics[width=0.3\linewidth]{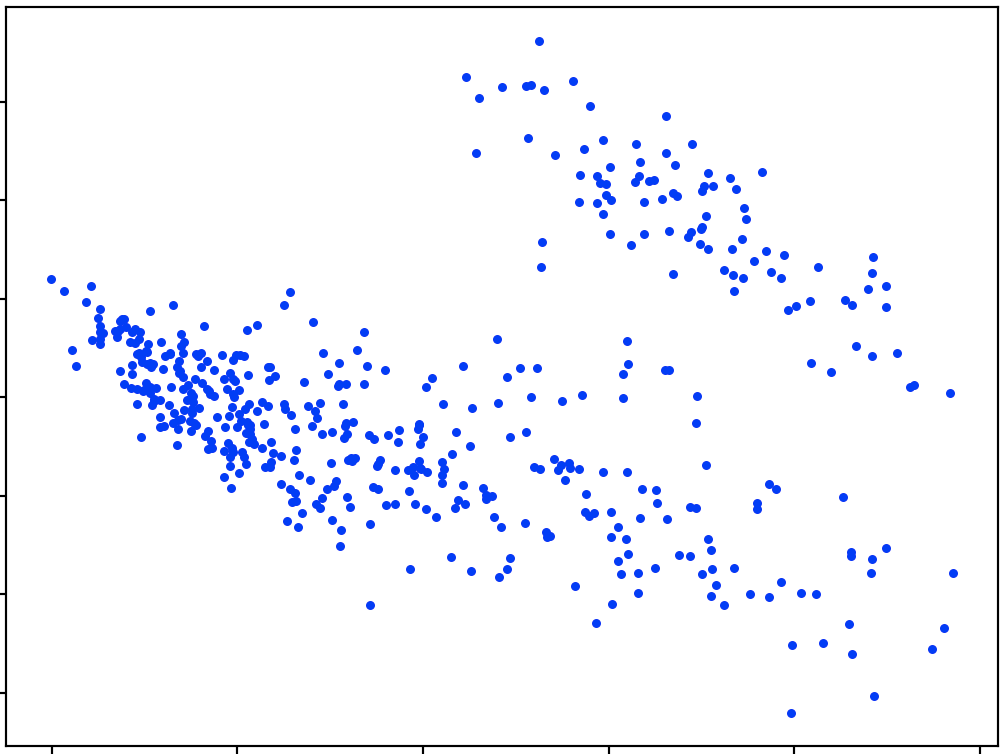}
    \includegraphics[width=0.3\linewidth]{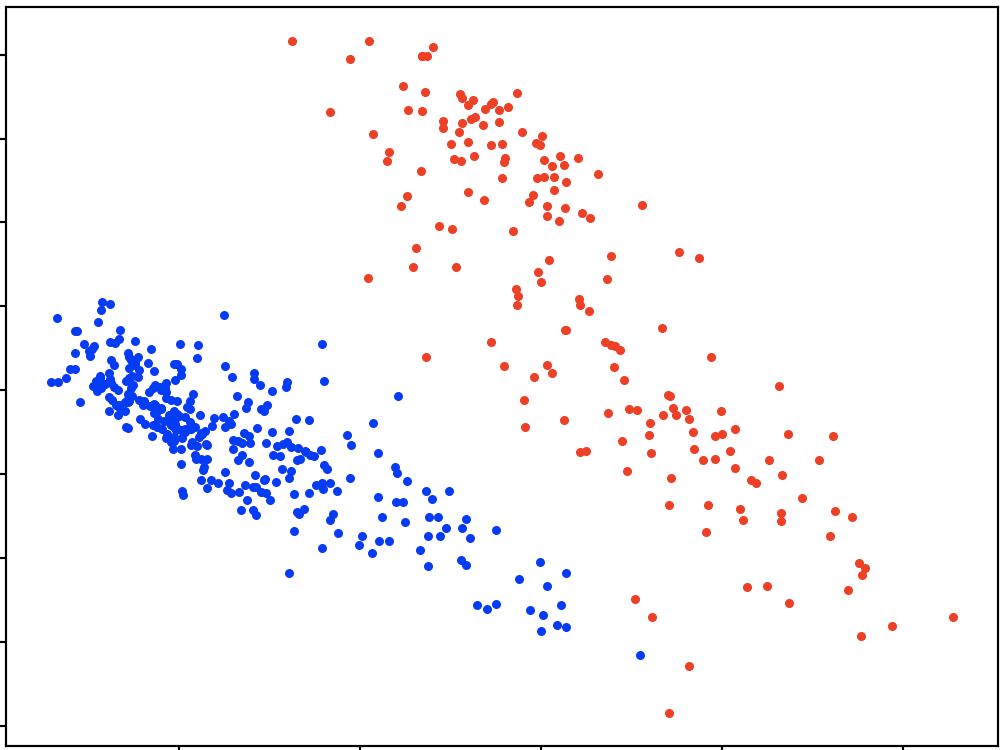}
    \includegraphics[width=0.3\linewidth]{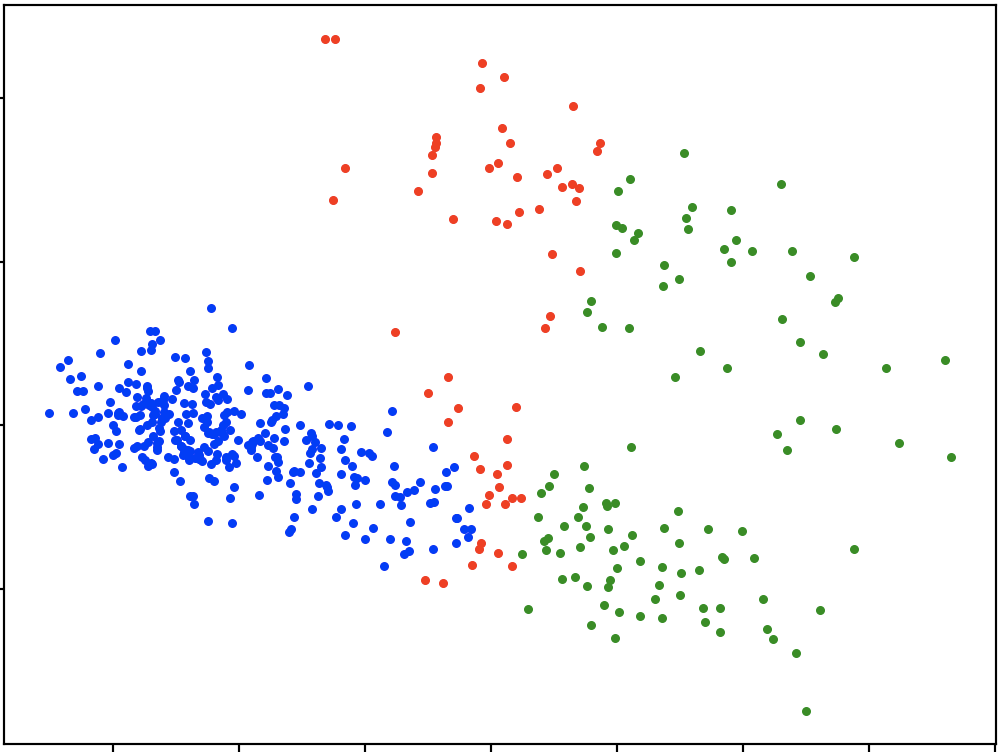}
    \caption{latent space clustering in Geo-query training data. The left figure shows how the training samples are projected on the latent space. The middle and right figures show the clustering results for two Gaussian mixtures and three Gaussian mixtures, respectively.}
    \label{fig:geo}
\end{figure}

The results also show the relationship between the performance and the number of filters, as well as the optimal number of filters.
On the Geo-query dataset, two-filter LEMS achieves the best performance. We can observe a negative relationship between the performance and the number of filters. While the number of filters goes up, both accuracies decrease. Figure \ref{fig:geo} shows the latent space representations for the training samples.

As we can see, once the number of filters goes up, the Gaussian mixture model $G$ is hard to give good clustering results. $G$ may make more mistakes in assigning data into clusters. This leads to the consequence that the data is not decoded by the filter that fits its features. Hence the performance will be deprecated if we choose the wrong number of filters.

\paragraph{Examples of Parsing Result}
\begin{itemize}
    \item ``what is the lonest river flowing through new york ?" = answer( longest( river( traverse( const( (stateID ( ``new york" ) ) ) ) ) )
    \item ``what is the lowest point in california ?" = answer( lowest( place( const( stateID( ``california" ) ) ) ) )
    \item ``tell me about the population of Missouri" = answer( population( const( stateID( ``Missouri" ) ) ) )
    \item ``name the rivers in Arkansas" = answer( river( loc( const( stateID( ``Arkansas" ) ) ) ) )
    \item ``give me the lakes in California" = answer( lake( loc( const( stateID( ``California" ) ) ) ) )
\end{itemize}
These sample results are generated from the 2-filter LEMS, which we get the best performance on.

In addition, to demonstrate why two filters perform the best, we select and display some examples from the two clusters.
\paragraph{Heterogeniety in the Question Structure}
The samples from the first cluster are in declarative form:
\begin{itemize}
    \item tell me what is the highest point in the state of Oregon
    \item tell me the capital of Texas
    \item tell me about the population of Missouri
    \item name the rivers in Arkansas
    \item name the 50 capitals in the USA
    \item give me the lakes in California
\end{itemize}

The samples from the second cluster are in the question form:
\begin{itemize}
    \item what state has the sparsest population density
    \item what rivers flow through Missouri
    \item what is the population of Portland Maine
    \item what is the longest river in Mississippi
    \item what is the largest state
    \item what is the capital of Michigan
\end{itemize}
From these samples, we can observe that the grammatical structure of the sentence largely affect the sentence representations.

\subsection{Machine Translation}

We conduct the second set of experiments on the Multi30k English-French dataset \cite{multi30k}. This is a dataset for neural machine translation and contains 30,000 pairs of training sentences and 1,000 pairs of validation sentences. The network and training specifications are identical to the previous experiments, presented in Table \ref{tab:specification}.

We use BLEU (bilingual evaluation understudy) score \cite{papineni-etal-2002-bleu} to evaluate the machine translation models and set $N = 4$ to compute the N-gram precision.
BLEU score is a score between 0 to 100. 0 represents that there is not even a single words in from the prediction matches the correct answer and 100 represents the prediction is exactly same with the answer. Generally, a score of 40 in English-French translation would be considered as a reasonable result. We present our results in terms of BLEU scores in Table \ref{tab:2}, for which the scores are the average of five experiments.

\begin{table}[!htbp]
\caption{Performance of the LEMS on different numbers of filters and a comparison between the LEMS and several baseline models.}
\begin{center}
 \begin{tabular}{||c c||} 
 \hline
 Model & BLEU\\
 \hline
 Baseline (text-only NMT) & 44.3\\
 SHEF \_ShefClassProj\_C\cite{Elliott_2017} & 43.6 \\
 CUNI Neural Monkey Multimodel MT\_C\cite{NeuralMonkey:2017} & 49.9 \\
 MGMAE & 44.1 \\ 
 2-filter LEMS & 43.6 \\ 
 3-filter LEMS & 44.9 \\
 4-filter LEMS & 46.3 \\
 5-filter LEMS & 45.1 \\
 \hline
\end{tabular}
\end{center}
\label{tab:2}
\end{table}

There are some examples of our model predictions on the testing data.
\paragraph{Examples of Translation Result}
\begin{itemize}
    \item \textbf{Input}: A young man skateboards off a pink railing.
    
    \textbf{Prediction}: Un jeune homme fait du skateboard sur une rampe rose.
    
    \textbf{Expected}: Un jeune homme fait du skateboard sur une rampe rose.
    
    \item \textbf{Input}: A young boy jumping from a bunk bed on a smaller bed.
    
    \textbf{Prediction}: un jeune garcon sautant d\&apos; un lit superposé sur un lit.
    
    \textbf{Expected}: Un jeune garçon sautant d\&apos; un lit superposé sur un lit plus petit.
    
    \item \textbf{Input}: Two people wearing odd alien-like costumes, one blue and one purple, are standing in a road.
    
    \textbf{Prediction}: Deux personnes portant des costumes bizarres de type alien, un bleu et d\&apos un violent par une route.
    
    \textbf{Expected}: Deux personnes portant des costumes bizarres de type alien, un bleu et un violet, sont debout sur une route.
\end{itemize}

\begin{figure}[!ht]
    \centering
    \includegraphics[width=0.3\linewidth]{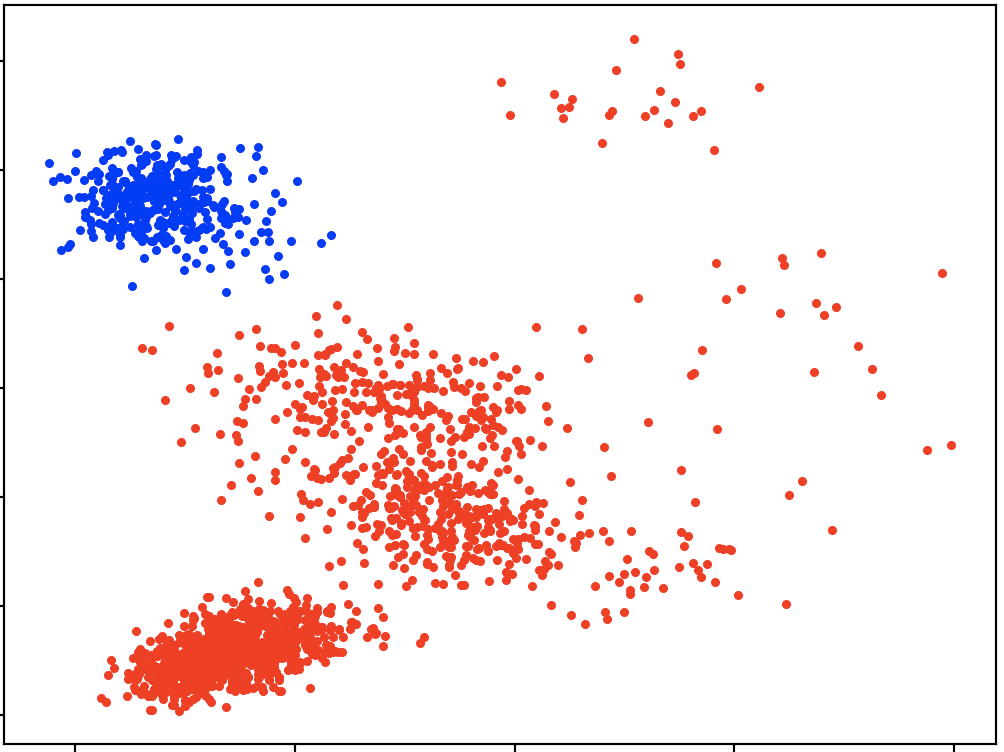}
    \includegraphics[width=0.3\linewidth]{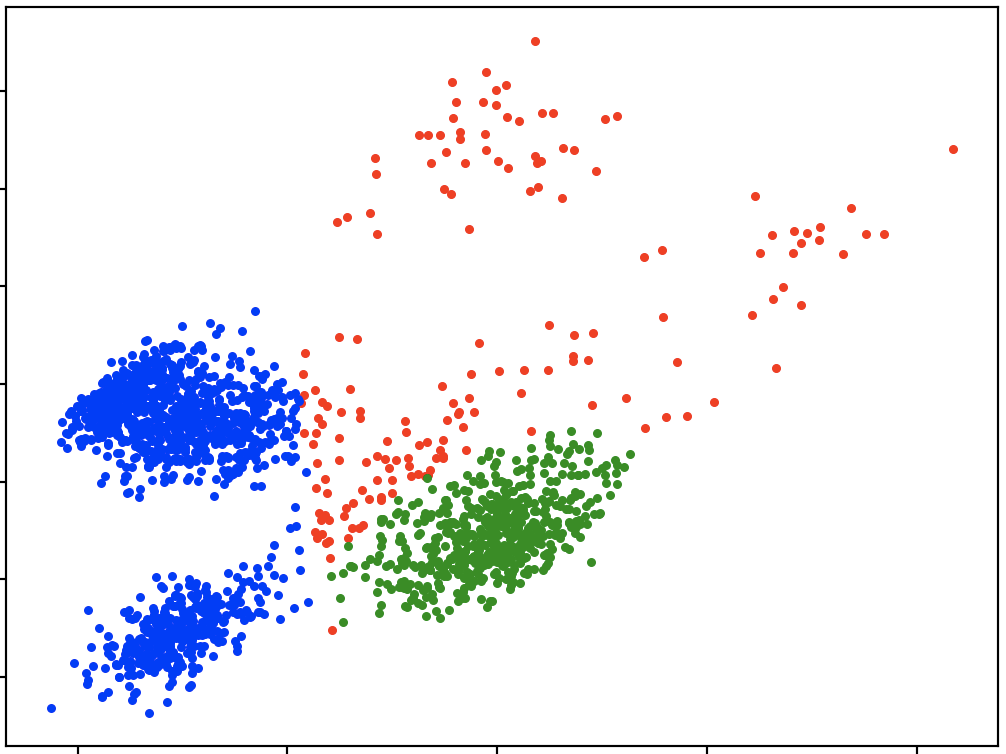}
    \includegraphics[width=0.3\linewidth]{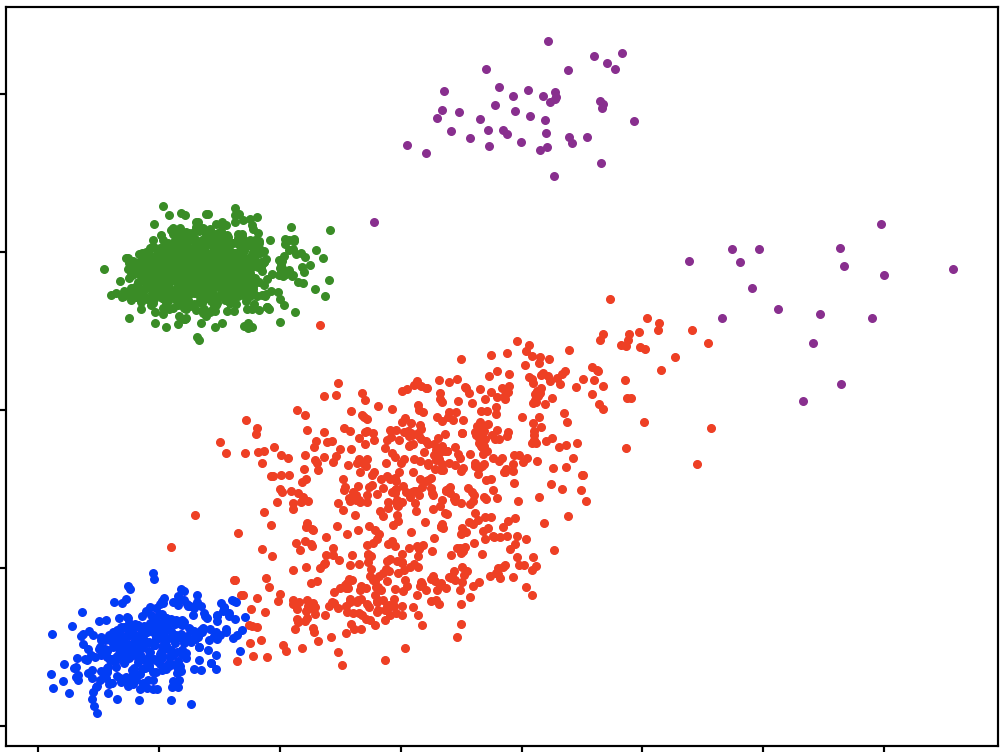}
    \caption{latent space clustering in Multi30k English-French translation training data. The figures from left to right present the latent space clustering results for two, three, and four Gaussian mixtures respectively.}
    \label{fig:trans}
\end{figure}

Table \ref{tab:2} indicates that our model can perform better than the ordinary encoder-decoder model, as well as some other baselines. By comparing with the MGMAE, we can observe a significant enhancement made by the transformer $T$. The transformer optimizes the latent space representations. Hence it leads to a improvement on constructing the outputs.
In this dataset, 4-filter LEMS achieves the best performance, the latent space can be visualized figure \ref{fig:trans}. Figure \ref{fig:trans} shows the clustering results under different numbers of clusters. Four clusters best separate the training data. Therefore, the filters associated with the four clusters can only concentrate on a similar set of features. Moreover, a better clustering model can reduce the probability of misclassified clusters in the testing stage, hence the performance of the model is ensured.

Then, we also want to explore why four filters work the best by inspecting the training samples. We select one training sample from each cluster from the 4-filter LEMS and display them below.
\paragraph{Heterogeniety in the Sentence Structures}
\begin{itemize}
    \item A girl plays in a small pool.
    \item A red-haired man with dreadlocks is sitting playing and acoustic guitar.
    \item A man that appears to be running in a marathon gives us 2 thumbs up.
    \item Riot police are standing in the background while a young man with a red scarf covering his face is walking.
\end{itemize}

We can observe the different grammatical structures of these samples and conclude that the grammatical structures of sentences significantly impact the clustering results. Specifically, the first sentence consists of a simple sentence structure, without any clauses, and in the present tense. The second sentence is in present continuous tense, and there is an attribute in front of the subject. In the third sentence, there is a clause for describing the subject. The last sentence consists of an adverbial clause which itself could be a complete sentence.

%% file: tex/discussion.tex
\section{Discussion and Future Direction}
\label{sec:discussion}
Experiments show that our LEMS is able to improve the performance on sequence-to-sequence tasks compared with the ordinary encoder-decoder model and MGMAE. One of the key components that significantly affect performance is the number of filters, i.e. the number of clusters in the latent space. To further explore the relationship between clustering quality and model performance, we calculated silhouette coefficients in each case. Each experiment was performed 5 times, and the average value is shown in Table \ref{tab:3}.

\begin{table*}[!htbp]
\caption{Silhouette Coefficient vs. model's performance in Geo-query dataset and Multi30k (English to French) dataset.}
\begin{center}
 \begin{tabular}{||c c c c c c||} 
 \hline
 Filter Number \vline & Geo Sil \vline & Geo Token \vline & Geo Denot \vline &  En-Fr Sil \vline & En-Fr BLEU\\
 \hline
 2 & 44.2 & 78.3 & 50.9 & 25.2 & 43.6 \\
 3 & 33.5 & 77.2 & 47.1 & 24.1 & 44.9 \\
 4 & 25.2 & 74.8 & 44.6 & 26.5 & 46.3 \\
 5 & 21.3 & 71.9 & 40.3 & 24.7 & 45.1 \\
 6 & 18.8 & 69.3 & 34.1 & 23.4 & 41.7 \\
 \hline
\end{tabular}
\end{center}
\label{tab:3}
\end{table*}

To determine the optimal number of mixtures for the GMM, we can use Silhouette Coefficient to evaluate the clustering performance given a specific number of mixtures, then choose the cluster number associate with the highest Silhouette Coefficient.

Table \ref{tab:3} shows how the silhouette coefficient affects the performance of the model. A positive correlation can be observed between the silhouette coefficient and each of the three quality metrics: token-level accuracy, representation accuracy, and BLEU score. This indicates that the quality of Gaussian mixture clustering is positively correlated with the prediction accuracy of the model. By choosing an appropriate number of clusters, better clustering results can be obtained. Therefore, we can get a set of data that is clearly separable under this number of clusters. Each filter can then be trained using a fixed set of unique features with less overlap with features in other clusters. Therefore, the filters can construct better outcomes.

\subsection{Future Research}

We are currently focusing on using Gaussian mixture models to divide the latent space. As a future direction, we can explore the potential use of other distributions such as Exponential distribution\cite{exponential} or Poisson distribution\cite{poisson}. There exists some scenarios that the data are not normally distributed. Furthermore, we can try to first learn the data distribution rather than assume they are normally distributed.

Another potential improvement is assigning cluster based on the output of the transformer. An addition linear layer can be added to the transformer to transform the output from hidden dimension to $n$-dimension, where $n$ is the number of clusters. Then, a softmax layer is added at the end of the transformer. At this stage, the output of the transformer becomes a probability vector $v_p$ that consists of the probabilities of assigning to each cluster. Hence if $v_p$ is the probability vector of sample representation $h_x$, then
\begin{center}
    $v_p[i] = P[h_x \text{ belongs to } i^{th} \text{cluster}]$.
\end{center}
We can take the cluster $c_j$ with the highest probability as the cluster the sample representation $x$ belongs to:
\begin{center}
    $h_x \in c_j \text{, where } j = argmax(v_p)$.
\end{center}
Then, the representation $h_x$ can be fed into the filter that corresponding to $c_j$. Hence the output can be constructed.

Furthermore, we have shown the positive relationship between the Silhouette Coefficient of the clustering result and the output quality. So, we can introduce a reinforcement learning model that using the Silhouette Coefficient as the reward to optimize the result of the model. For instance, an Advantage Actor Critic (A2C)\cite{mnih2016asynchronous} reinforcement learning algorithm can be utilized for optimizing the trainable parameters in the transformer.

The reward function is based on the Silhouette Coefficient $S_c$:
\begin{equation}
    r = a \cdot S_c + b
    \label{eq:reward}
\end{equation}
where $a$ and $b$ are the user-defined constants. To maximize the reward, the RL model learns actions that adjust the trainable weights in the transformer $T$. The action function is defined as:
\begin{equation}
    T_i = \Vec{a} \cdot T_i'
    \label{eq:action}
\end{equation}
where $\Vec{a}$ is the action, $T_i'$ is the set of old parameters of the $i^{th}$ layer of $T$, $T_i$ is the set of updated parameters of the $i^{th}$ layer of $T$.

A target Silhouette Coefficient $S_c^{target}$ can be set as the terminate state of the RL model. The learning process immediately stop once the Silhouette Coefficient $S_c$ reaches the target score: $S_c \ge S _c^{target}$. This reinforcement learning model is expected to enhance the final results by improving the quality of clustering.

%% file: tex/conclusion.tex
\section{Conclusion}
We introduced the Latent Enhanced Multi-filter Seq2Seq model for sequence to sequence tasks. Our model addresses data heterogeneity by dividing the sample into relatively homogeneous groups and utilizing multiple filters so that each filter corresponds to a group. We demonstrate the importance of the multi-filter architecture by conducting different comparative experiments. Simultaneously, we show the heterogeneities of grammatical structures of the training samples. We also introduce the transformer for latent space enhancement. The transformer enhances the qualities of the representations, which also leads to an observable improvement.